\documentclass{article}
\usepackage{ijcai20}
\usepackage{color}
\usepackage{times}
\usepackage{soul}
\usepackage{url}
\usepackage[hidelinks]{hyperref}
\usepackage[utf8]{inputenc}
\usepackage[small]{caption}
\usepackage{graphicx}
\usepackage{amsmath}
\usepackage{amsthm}
\usepackage{booktabs}
\usepackage{algorithm}
\usepackage{algorithmic}
\usepackage{amsmath,amsfonts,amsthm,bm}
\title{Solving Hard AI Planning Instances Using Curriculum-Driven Deep Reinforcement Learning}

\author{
Dieqiao Feng \and
Carla P. Gomes \And
Bart Selman
\affiliations\
Department of Computer Science\\
Cornell University
\emails
\{dqfeng, gomes, selman\}@cs.cornell.edu
}
\begin{document}

\maketitle

\begin{abstract}
    % edits by Bart
    % We present a new approach to uncover structure in complex 
    % combinatorial search domains by tackling hard instances of AI % planning problems. We use Sokoban as our demonstration 
    % domain.
    
    % Sokoban is a notoriously hard planning problem. Current AI % planners can only solve relatively small instances.
    
     Despite significant progress in general AI planning, certain domains remain out of reach of current AI planning systems. Sokoban is a PSPACE-complete planning task and represents one of the hardest domains for current AI planners. Even domain-specific specialized search methods fail quickly due to the exponential search complexity on hard instances. Our approach based on deep reinforcement learning augmented with a curriculum-driven method is the first one to solve hard instances within one day of training while other modern solvers cannot solve these instances within any reasonable time limit. In contrast to prior efforts, which use carefully handcrafted pruning techniques, our approach automatically uncovers domain structure. Our results reveal that deep RL provides a promising framework for solving previously unsolved AI planning problems, provided a proper training curriculum can be devised.
    
    % We evaluate our approach on hard Sokoban instances and show 
    % that our approach facilitates learning complex planning 
    % invariants and demonstrates near polynomial running time 
    % growth on self-curriculum setting as the size of subcases
    % increases.
\end{abstract}

\section{Introduction}
Deterministic, fully observable planning is a key domain for artificial intelligence. In its full generality, AI planning encompasses general theorem proving, where proofs can be viewed as plans leading from a set of basic axioms to the theorems to be proved. Planning is well known to be a very challenging computational problem: finding proofs in a strong first-order mathematical theory which encodes basic arithmetic is undecidable \cite{godel1931formal} and plan-existence is PSPACE-complete for propositional STRIPS planning \cite{bylander1994computational}. Domain-independent planners, such as BLACKBOX \cite{kautz1998blackbox}, IPP \cite{koehler1997extending}, and FF \cite{hoffmann2001ff} among many others, were built for solving general planning tasks given by an initial state, goal state, and a set of plan operators \cite{vallati20152014}. Though these planners have greatly enlarged the set of feasible planning tasks, one major shortcoming of these planning systems is that they may do well on one problem domain but poorly on another, which has prevented a wider use of AI planning systems. This situation is in contrast to the development of SAT/SMT solvers, which also tackle a combinatorial search task, but have found wide applicability in, for example, hardware and software 
verification \cite{jarvisalo2012international}. An alternative approach to general AI planning is to develop domain-specialized solvers, e.g., Sokolution for solving Sokoban planning problems, as discussed below. The specialized solvers utilize handcrafted domain-specific knowledge to prune the search space. Clearly, an effective domain-independent approach is preferable. Our learning framework presented here provides a path towards such domain independence. In particular, we will use a machine learning framework to automatically uncover domain-specific problem structure during the solution process.

% Bart: took out lee2020learning
% Bart: hmm. Lee reference is perhaps for related
% work. Not good here. Soving QBF is not yet an
% an example of a successful approach. The state-of-the-art of
% QBF is as bad as FOL.

% Bart: Vampire prover \cite{kovacs2013first} for first-order 
% logic and
% Bart: I took this out. Most people would not view Vampire
% as a specialized solver since it tackles FOL.

Recent advances in the deep learning community inspired methods of augmenting search with deep neural networks using deep reinforcement learning (RL). In the game domain, AlphaGo \cite{silver2016mastering} as the first Go program to beat professional players in 2016 and its more general and newest version AlphaZero \cite{silver2017mastering} achieved a higher Elo rating and dominated the state-of-the-art Chess program Stockfish. One key question about the success of deep RL in these combinatorial (logical) domains is {\em whether the game setting is a required component for success.} RL requires a reward signal. In a game setting, this signal comes from the ultimate win/loss result from playout. For a game, we can train the deep nets in a self-play approach. In such an approach, the initial deep net starts off playing at a very low level (essentially random play). But in self-play against an equally weak player (using a copy of the trained network), the system will see a mixture of wins and losses and thus gets a useful reward signal about the utility of states. The system can then slowly improve its level of play through repeated rounds of self-play. 

The core challenge in a single-agent setting, such as AI planning, where we want to solve unsolved problem instances is: {\em how do we get any positive reward signal during training?} This is because a positive feedback signal would require a valid plan to the goal but that is exactly what we are looking for. In fact, the problem instances we will solve here require very subtle chains of several dozens to even hundreds of steps. A random exploration will never ``accidentally'' encounter a successful chain. Our solution is to devise a series of training instances (a ``curriculum'') that slowly builds up to the full, previously unsolved problem instance. We will see below how such an approach is general and surprisingly effective. Curriculum based training has earlier been proposed in \cite{bengio2009curriculum} as a strategy for partitioning training data for incremental training of hard concepts. A novel aspect of our setting is that at each level of our curriculum training, we use what was learned at the previous level to obtain new training data to reach the next level. 

Given the PSPACE-completeness of many interesting planning tasks, it is widely assumed (unless P = PSPACE) that developing a solver capable of solving effectively any arbitrary instance is infeasible. Moreover, due to the significant overhead of training deep neural networks, we do not aim to compete on running time with finely tuned specialized solvers on small problem instances. In this work, we therefore focus on planning instances that are right beyond the reach of current state-of-the-art specialized solvers. We will show for the first time how such instances can be tackled successfully within a deep RL framework. Specifically, we will show AI planning instances on which our deep learning strategy outperforms the best previous combinatorial search methods. We will also provide insights about what problem structure deep nets capture during the learning process.

We selected Sokoban planning as our AI planning task because of its extreme difficulty for AI planners \cite{fern2011first} \cite{lipovetzky2013structure}. Moreover, Sokoban instances have a regular 2-D input shape that is well-suited for convolutional neural networks. Such 2-D structure can also be found in many other AI planning that involve scheduling and transportation style problems. However, Sokoban is much more challenging in computational terms. Sokoban is a single-player game, created in 1981, in which, given a set of boxes and equal number of goal locations, a player needs to push all boxes to goal squares without crossing walls and boxes. Figure~\ref{fig:sokoban} shows a typical instance. The player can only move horizontally or vertically onto empty squares. Despite its apparent conceptual simplicity, it quickly became clear that one could create very hard instances with highly intricate and long solutions (if solvable at all). Analyzing the computational complexity of Sokoban is non-trivial but the question was finally resolved by Culberson in 1997, who proved the problem to be PSPACE-complete \cite{culberson1997sokoban} \cite{hearn2005pspace}. We will show below that the harder Sokoban instances lie far beyond general AI planners but also quickly are beyond the reach of specialized Sokoban solvers. All modern state-of-the-art solvers are based on a combinatorial search framework augmented with intricate handcrafted pruning rules and dead-end detection techniques.

Our framework learns and solves a single hard Sokoban instance at a time. This is an important choice in our setting. We want {\em the deep net to uncover the underlying structure of the combinatorial space} that is directly relevant to the hard --- previously unsolved --- instance under consideration. This approach mimics conflict-driven clause learning (CDCL) \cite{marques1999grasp} for solving Boolean satisfiability problem (SAT). In SAT solving, the clauses are learned {\em during the processing of a single instance.} In this setting, the learned clauses are optimally relevant to the problem instance at hand. Another potential advantage of our framework is that all the parameters of the deep neural network are focused on the layout of the given input instance and its corresponding search space. Though some general knowledge about Sokoban, e.g., that pushing a box to a corner leads to a dead-end state, can be learned from one instance and generalized to others, we show that our training setup can also discover this kind of knowledge efficiently and generalize well across its search space. In addition, the deep learning framework can now uncover very specialized problem structure that helps tailor the search for the solution to the specific problem instance at hand. Examples of such structure can be a certain placement of a subset of boxes from which the goal state cannot be reached. The search mechanism can now eliminate any exploration action sequences that lead to such a placement. This is analogous to the pruning provided by learned clauses in SAT solvers. The learned clauses are specific to the SAT instance under consideration.

As we discussed above, we need to devise a way to obtain a proper training signal for solving AI planning problems. Since the input instance might be extremely hard and therefore cannot directly provide any positive reward signal, we incorporate the idea of \textit{curriculum learning} and construct simpler subcases derived from the original challenge problem. In our Sokoban domain, a natural choice is to randomly select smaller subsets of initial boxes and goal squares while leaving all walls unchanged. In particular, our learning procedure starts from exploring 2-box subcases and gradually increases the number of boxes after the success rate of finding a solution increases to a certain threshold. We show that knowledge learned from subcases with smaller numbers of boxes can generalize successfully to subcases with larger numbers of boxes.

% Bart: I cut beginning because RL and MCTS are actually 
% generally used
% for dealing with incomplete, noisy, and probabilistic states.
% It's their use in a complete deterministic full information 
% setting that is somewhat unusual.
%
% Unlike many real-world planning tasks, deterministic, fully 
% observable planning tasks have properties of having a perfect 
% model and a strong simulator, 
Solving the Sokoban planning task is
a combinatorial search problem and we will utilize AlphaZero-style Monte Carlo tree search in reinforcement learning for exploring the search space more efficiently. The only domain knowledge we use during learning is computing valid pushes from a state and building the state transition table, and the input of the neural network is the current raw 2-D board state. Intricate handcrafted techniques in modern solvers like dead-end detection are not used.

Our experiments reveal that our curriculum-driven deep reinforcement learning framework can surpass traditional specialized solvers for a large set of instances from benchmark datasets such as XSokoban and Sasquatch. The deep network helps the Monte Carlo tree search explore the search space more effectively and offers significant generalization to unseen states. In addition, the growth of running time when the complexity of the instances in the curriculum increases is near polynomial instead of exponential.

We will also provide a number of other insights into the learning process. Of particular interest is the observation that when training the deep net on harder tasks in the curriculum, its performance on easier instances degrades. This form of ''catastrophic forgetting'' makes the stronger networks less robust, even when better at solving harder instances. It would be an interesting research direction to devise a curriculum-driven approach that does not show degradation on easier tasks while still reaching maximal effectiveness on the original problem.

\section{Related Work}
Search in combinatorial domains has been studied extensively in AI, in areas such as planning, decision making, and reasoning. For NP-complete tasks, successful SAT solvers WalkSAT \cite{selman1992new} and the CDCL framework \cite{marques1999grasp} have been built to efficiently uncover structure of the input problem and demonstrate near polynomial scaling on many industrial SAT domains. The key insight of their success is the ability of the algorithm to learn problem invariants and reshape the search space by avoiding entering subtrees which do not contain a solution. Most planning tasks are harder than SAT and usually are at least PSPACE-complete. Graphplan \cite{blum1997fast} and FF are general planners accepting formal languages such as PDDL \cite{fox2003pddl2}. However, as reported in \cite{welle2003sokoban}, a major shortcoming of these general planners is that they may do well on one problem domain but poorly on another.

The enhancement of planning with learning \cite{fern2011first} has been investigated extensively in the past. Directed by the current goal, \cite{abel2015goal} prune away irrelevant actions. In each state, \cite{rosman2012good} exploit the usefulness of each action by learning action priors. For Sokoban-specialized solving, modern solvers utilize intricate domain-dependent techniques such as subtle dead-end detection, duplicate positions pruning, lower bound calculation, and no influence move detection \cite{junghanns2001sokoban}. While all of these techniques offer efficiency improvements over general planning for Sokoban, good representations of states, tight heuristic functions as well as dead-end detection methods are handcrafted, which requires a careful inspection of domain structure and heavy utilization of domain knowledge.

In recent years, deep neural networks have achieved promising results in many domains. The most exciting result in the combinatorial domain is AlphaZero which utilizes deep reinforcement learning to automatically discover domain structure of two-player games like Chess and Go. Key to its success is the self-play learning strategy, which starts with two weak players and gradually collects useful learning signals by self-play and gradually improves the ability of the players. Previously, it was not clear how to develop such ``curriculum-driven'' strategy in the planning domain. Because unlike in the game domain, where learning signals (wins/losses) are available for any pair of players of roughly equal strengths, including very weak and random players, in the planning domain, the agent will initially fail to reach the goal state at every attempt and therefore cannot bootstrap its learning process.

Deep neural networks have also been used to help tackle Sokoban problems. \cite{weber2017imagination} augment deep reinforcement learning with an imagination component, and \cite{groshev2018learning} use imitation learning to learn from successful Sokoban plays and generalize reactive policies to unseen instances. However, their performance is nowhere close to state-of-the-art specialized Sokoban solvers.

Our approach offers two major advantages over prior approaches: (1) our approach solves hard benchmark instances that are out of reach of specialized Sokoban solvers; (2) no domain-specific knowledge is needed during learning and no extra data, e.g., manually provided solutions, are required. Our idea of training on similar, but easier subcases out from the original instance can be adapted to other planning domains.

\section{Formal Framework}
\subsection{Model}
\begin{figure}[t]
    \centering
    \begin{minipage}{0.5\linewidth}
        \centering
        \includegraphics[width=\linewidth]{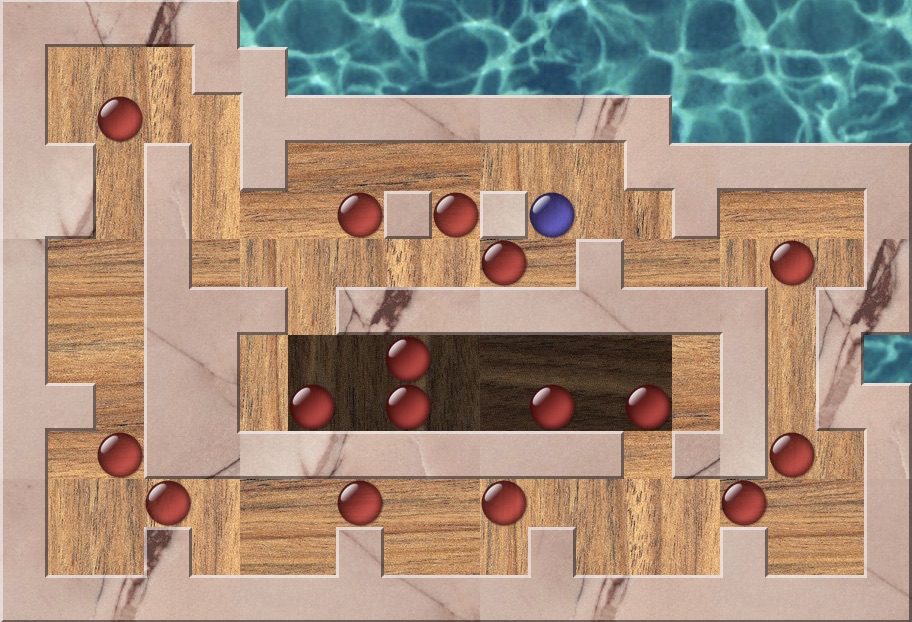}
    \end{minipage}%
    \begin{minipage}{0.5\linewidth}
        \centering
        \includegraphics[width=\linewidth]{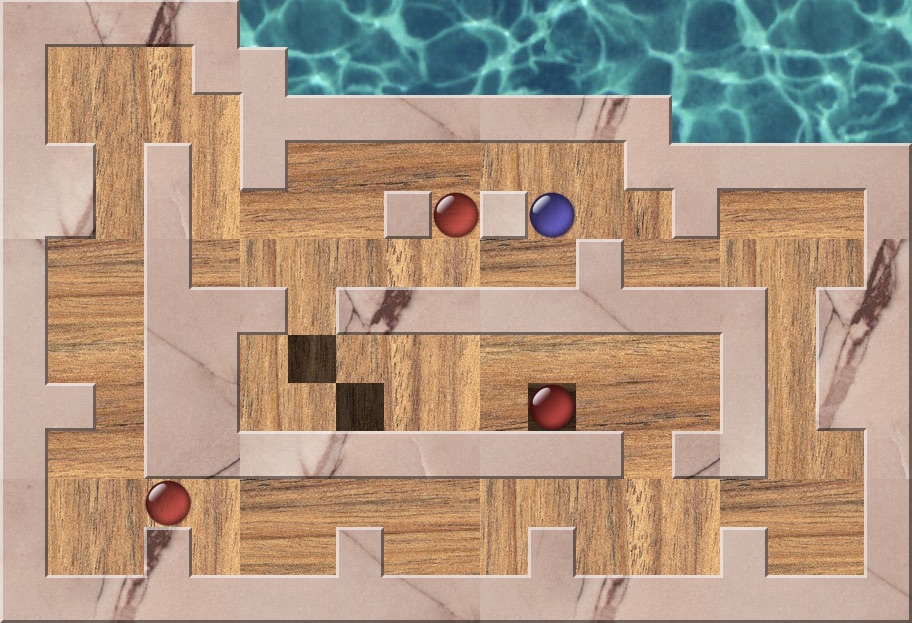}
    \end{minipage}
    \caption{The instance XSokoban\_29 (left panel) and  one of its 3-box subcase (right panel). The blue circle is the location of the player (or ``pusher''), red circles are boxes, and cells with dark background are goal squares. Light colored squares form walls. The player has to push the boxes onto goal squares.
    }
    \label{fig:sokoban}
\end{figure}
Given a Sokoban instance $\mathcal{I}$, the preprocessing phase computes the set of all possible pushes $\mathcal{A}$. A deep neural network $(\bm{p}, v) = f_\theta(s)$ with parameters $\theta$ takes the board state $s$ as input and outputs a vector of action probability $\bm{p}$ with components $\bm{p}_a = \mathrm{Pr}(a|s)$ for each push action $a\in\mathcal{A}$, and a scalar value $v$ indicating the estimated number of remaining steps to the goal from state $s$. The left figure of Figure \ref{fig:sokoban} shows the original instance XSokoban\_29 from the benchmark dataset XSokoban. The input to the network is a $6\times H\times W$ image stack consisting of 6 features planes while $H$ and $W$ are the height and width of the corresponding Sokoban instance. Feature planes represent walls, empty goal squares, boxes on empty squares, boxes on goal squares, player-reachable cells, and player-reachable cells on goal squares respectively.

The effort of solving a Sokoban instance can be divided into two parts:
\begin{enumerate}
    \item The player moves to the correct position adjacent to a box for pushing.
    \item The player pushes the box.
\end{enumerate}
In our experiment, we use the set of valid pushes instead of valid moves as the action set since the number of pushes in a solution is significantly smaller than the number of moves. (One move is moving one square over for the player (the pusher).) In other words, the plan length is generally far shorter in terms of number of pushes vs.\ number of moves. To model the set of valid pushes at each state requires keeping track of the reachable cells that are next to blocks for the player. Illegal pushes are masked out by setting their probabilities to zero, and re-normalising the probabilities for remaining moves.
% The number of valid pushes is linear in the number of boxes.
% The number of valid moves is at most four: move upward, rightward, downward, and leftward, while the size of valid pushes is linear to the number of boxes. The set of valid pushes 
% requires computing reachable cells from the player which involves extra domain knowledge.

For each instance $\mathcal{I}$, we set a maximum allowable pushes or ``steps'' $\mathcal{I}_{\mathrm{max}}$ during the learning phase. $\mathcal{I}_\mathrm{max}$ indicates the maximum number of steps of a single plan that the algorithm is allowed to explore during learning. The model will be forced to stop after $\mathcal{I}_\mathrm{max}$ pushes. Setting such a threshold can help avoid infinite meaningless loops when exploring. The remaining-step estimator $v$ is also  normalized to the interval $[0, 1]$ to fit better into the neural network framework. Notice that if $\mathcal{I}_\mathrm{max}$ is set smaller than the length of the shortest solution plan then the model will never find any solution. In our experiments, we start with $\mathcal{I}_\mathrm{max} = 500$ and double it whenever learning fails after a long run.

The learning framework consists of multiple iterations, and each iteration contains three parts:
\begin{enumerate}
    \item Initial board generation phase: we randomly generate 500 initial boards according to our curriculum-driven strategy described in subsection \ref{curriculum}.
    \item Exploration phase: the model searches for solutions (plans) for these boards with Monte Carlo tree search (MCTS) driven by the policy/value network trained so far. For more details see subsection \ref{mcts}.
    \item Training phase: we train the neural work with learning signals collected from the exploration phase. This part will be further illustrated in subsection \ref{train}.
\end{enumerate}

\subsection{Curriculum-driven Strategy}
\label{curriculum}
For hard Sokoban instances the deep RL setup may fail to find any solution and thus gets no useful training signal. Our curriculum strategy is based on two insights: 
(1) construct simpler subcases that are more likely to be solved by the current trained model; 
(2) the constructed subcases should share similar structure information with the original instance to enhance knowledge generalization from a series of subcases to the original problem instance.

Learning starts by choosing a small subset of initial boxes and goal squares to form a subcase for exploration and training while leaving wall locations unchanged. Figure \ref{fig:sokoban} (right) shows one such example. Three boxes and goal squares are randomly selected from the initial ones. The resulting subcase requires a much simpler plan. Specifically, assume the input Sokoban instance has $n$ boxes and goal squares, in each iteration we randomly select $m \leq n$ boxes and goal squares and gradually increase $m$ after a certain level of performance has been reached at each level. Compared with the original problem, solutions of the subcases will be shorter and, most importantly, easier to find with MCTS and the deep net trained so far. By solving a collection of $m$ box subcases for each value of $m$, we effectively train a distance function and action model that can handle $m$ box subproblems for a range of initial and goal placements. This level of generality is important because we do not know in advance on which goal square, any particular box from the initial state will end up. Moreover, because we start with a 2 box subcase, MCTS with a randomly initialized deep net can still find a solution path, and thus a positive reward signal. By slowly increasing the subcase size (and difficulty), deep RL can continue to obtain a positive reward signal and incrementally improve the trained net to handle increasingly complex scenarios, ultimately leading to a solution to to the original instance, when $m = n$.

In the experiment section we will show that it is necessary for the model to learn to a certain accuracy rate on $m$-box subcases before jumping to $(m+1)$-box subcases. Specifically, if the model jumps to $(m+1)$-box scenarios before it reaches a high success rate on $m$-box scenarios, one potential danger is that the performance of learning will abruptly degrade and the model might be no longer able to find any solution for $(m+1)$-box subcases.

\begin{figure}[t]
    \centering
    \begin{minipage}{0.5\linewidth}
        \centering
        \includegraphics[width=\linewidth]{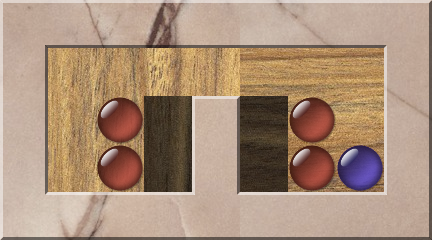}
    \end{minipage}%
    \begin{minipage}{0.5\linewidth}
        \centering
        \includegraphics[width=\linewidth]{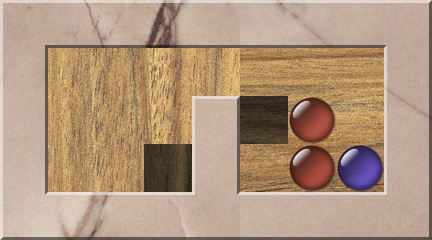}
    \end{minipage}
    \caption{One example showing that subcases are not necessarily solvable even though the original instance has a solution.
    }
    \label{fig:small_sokoban}
\end{figure}
To decide on when to increase $m$, one possible measure to use would be the solution rate reached at that level. However, somewhat counterintuitively, even if the original problem instance is solvable, certain subcases may not have solutions. This is due to a hidden complexity of the Sokoban domain: we know that the boxes need to reach the goal squares but we don't know exactly which box should go to which goal square. So, even though our subproblems use a strict subset of the boxes and goal squares, we may accidentally generate an unsolvable subproblem. Figure \ref{fig:small_sokoban} gives an example. In this case, the probability of generating a solvable 2-box subcase is only $\frac{1}{2}$, since we need to guarantee that the number of boxes and goal squares be the same in each room.
% Bart: Dieqiao please check. I get 18/36. Not 11/36.
In general, it is difficulty to compute the probability that a random subcase is solvable. Therefore, using the success rate at a certain level is not a robust criterion. We use an alternative way to decide when to increment $m$. Specifically, we increment $m$ when the success rate of finding a solution has not improved over a certain number of iterations. We use 5 iterations in our experiments. Our experiments show that this strategy works well in practice.
% This method has a potential issue that the model might jump to $m+1$ when the success rate sticks in a local minima. Our experiments show that this criterion works in most instances we have tested.

\subsection{Monte Carlo Tree Search}
\label{mcts}
We search for  solutions (plans) of the Sokoban m-box subinstances ($m \leq n$) using an AlphaZero-style Monte Carlo tree search (MTCS) guided by the deep net trained so far. As we will see, MTCS works well but other search techniques may also be worth exploring in future work. In MTCS, at each state $s$, we compute $(\bm{p}, v) = f_\theta(s)$ and create a root node $R$ which contains the state $s$. Multiple Monte Carlo rounds will be performed from $R$ to calculate the best move in $s$. Each round consists of three components:
\begin{itemize}
    \item Selection: start from the root node $R$, which contains the state $s$, and select successive child nodes which maximize a utility function until a leaf node $L$, the goal, or a dead-end is reached. A leaf node is any node that has never been evaluated by the neural network before. If the goal or a dead-end is reached then we jump to the backpropagation phase otherwise the expansion phase.
    \item Expansion: compute the set of all valid pushes $\mathcal{A}$ from the state $s_L$ of the node $L$. Unlike traditional MCTS followed by a roll-out which simulates multiple random plays, we evaluate $(\bm{p}, v) = f_\theta(s_L)$ with the neural network and use $v$ as the estimated evaluation for the backpropagation phase.
    \item Backpropagation: use $v$ of $s_L$ to update information of the nodes on the path from $R$ to $L$. Set $v$ to $0$ if $s_L$ is the goal state or $1$ if $s_L$ is a dead-end. Assume the state-observation-action trajectory from $R$ to $L$ is $s = s_0 \xrightarrow{a_1} s_1 \xrightarrow{a_2} \cdots \xrightarrow{a_l} s_l = s_L$ where $l$ is the length of the trajectory, we update $Q_\mathrm{new}(s_i, a_{i+1})$ to
    \[
    \frac{Q(s_i, a_{i+1}) \cdot N(s_i, a_{i+1}) + \min(v + \frac{l-i}{\mathcal{I}_\mathrm{max}}, 1)}{N(s_i, a_{i+1}) + 1},
    \]
    where $Q(s, a)$ is the mean action value averaged from previous rounds and $N(s, a)$ is the visit count.
\end{itemize}
To select child nodes, we choose $a_t=\mathrm{argmax}_a Q(s_{t-1},a)+U(s_{t-1},a)$ using a variant of the PUCT algorithm where
\[
    U(s,a) = \mathrm{cput}\cdot\frac{\sqrt{1 + \sum_b N(s, b)}}{1 + N(s,a)}\cdot \bm{p}_a,
\]
and $\mathrm{cput}$ is a constant balancing exploration and exploitation.

After 1600 rounds have been performed, we choose a move either greedily or proportionally with respect to the visit count at the root state $s$. This procedure proceeds until a dead-end or the goal is reached, or the maximum allowable pushes $\mathcal{I}_\mathrm{max}$ have been performed. Notice we don't utilize any advanced dead-end detection algorithm used in previous modern solvers. Instead, we only detect dead-ends when no valid pushes from the state are available, e.g., all boxes are pushed into corners and are no longer movable.

To construct learning signals for the training phase, we collect all states on paths explored by the Monte Carlo tree search, and use the probability proportional to the visit count as the improved probability distribution $\bm{\pi}$ for training. For value prediction, if the leaf node is the goal state then we use the distance to the leaf node as the new label $u$. Otherwise, either a loop or a dead-end is reached and we set $u$ to 1 for all nodes on the path.

\subsection{Training}
\label{train}
We use 5 GPUs to train the network and each iteration contains 1000 epochs with mini-batch 160 in total. Unlike \cite{mnih2013playing} and \cite{silver2017mastering} who maintain an extra memory pool to save training episodes, we directly train on data collected from the current iteration. Specifically, the network parameters $\theta$ are adjusted by gradient descent on the loss function that sums over a mean-squared loss and a cross-entropy loss
\[
l = (u - v)^2 - \bm{\pi}\log(\bm{p}) + c\cdot \|\theta\|^2,
\]
where $c$ is the constant to control the impact of weight decay. After the training phase, new parameters of the network are used to guide the Monte Carlo tree search in the next iteration.

For this paper, we did not perform a detailed hyper parameter study to select the best network structure for our problem setting. We used vanilla ResNet \cite{he2016deep} with 8 residual blocks as the network setting for all experiments. It is an indication of the promise of our general framework that we already obtained good results with standard ResNet. The overall performance can likely be further improved with careful hyper parameter tuning.

\section{Experiments}
Here we report our experiments on XSokoban, the de facto standard test suite in the academic literature on Sokoban solver programming, as well as other large test suites\footnote{Sokoban datasets available at \url{http://sokobano.de/wiki/index.php?title=Solver_Statistics}}. We pick instances that are marked as "Solved by none of the four modern solvers". The time limit for the statistics of previous benchmark is usually 10 minutes. We extend the time limit to 24 hours if the option is available and retest all solvers on these hard instances. We do the test on our method, state-of-the-art Sokoban-specialized solver Sokolution, and domain-independent general planner FF.
Table \ref{tab:table1} shows the performance of each solver on the selected instances.

\begin{table}
\centering
\begin{tabular}{lrrr}
\toprule
Sokoban instance  & Our method & Sokolution & FF \\
\midrule
XSokoban\_29 & 9.1h & Failed & Failed \\
Sasquatch\_29 & Failed & Failed & Failed \\
Sasquatch\_30 & Failed & Failed & Failed \\
Sasquatch3\_18 & 14.9h & Failed & Failed \\
Sasquatch7\_48 & 23.4h & 1.0h & Failed \\
Grigr2001\_2 & 22.1h & Failed & Failed \\
\bottomrule
\end{tabular}
\caption{Performance comparison. Sokoban instances are selected from standard datasets and are marked as "Solved by none". Time limit for all solvers are extended to 24 hours if the option is available. All solvers are running on the same CPU cores while our method utilizes additional 5 GPUs. The conversion from Sokoban into STRIPS format is shown in \protect\cite{welle2003sokoban}.}
\label{tab:table1}
\end{table}

\subsection{Scaling Comparison}
\begin{figure}[t]
    \centering
    \includegraphics[width=1.0\linewidth]{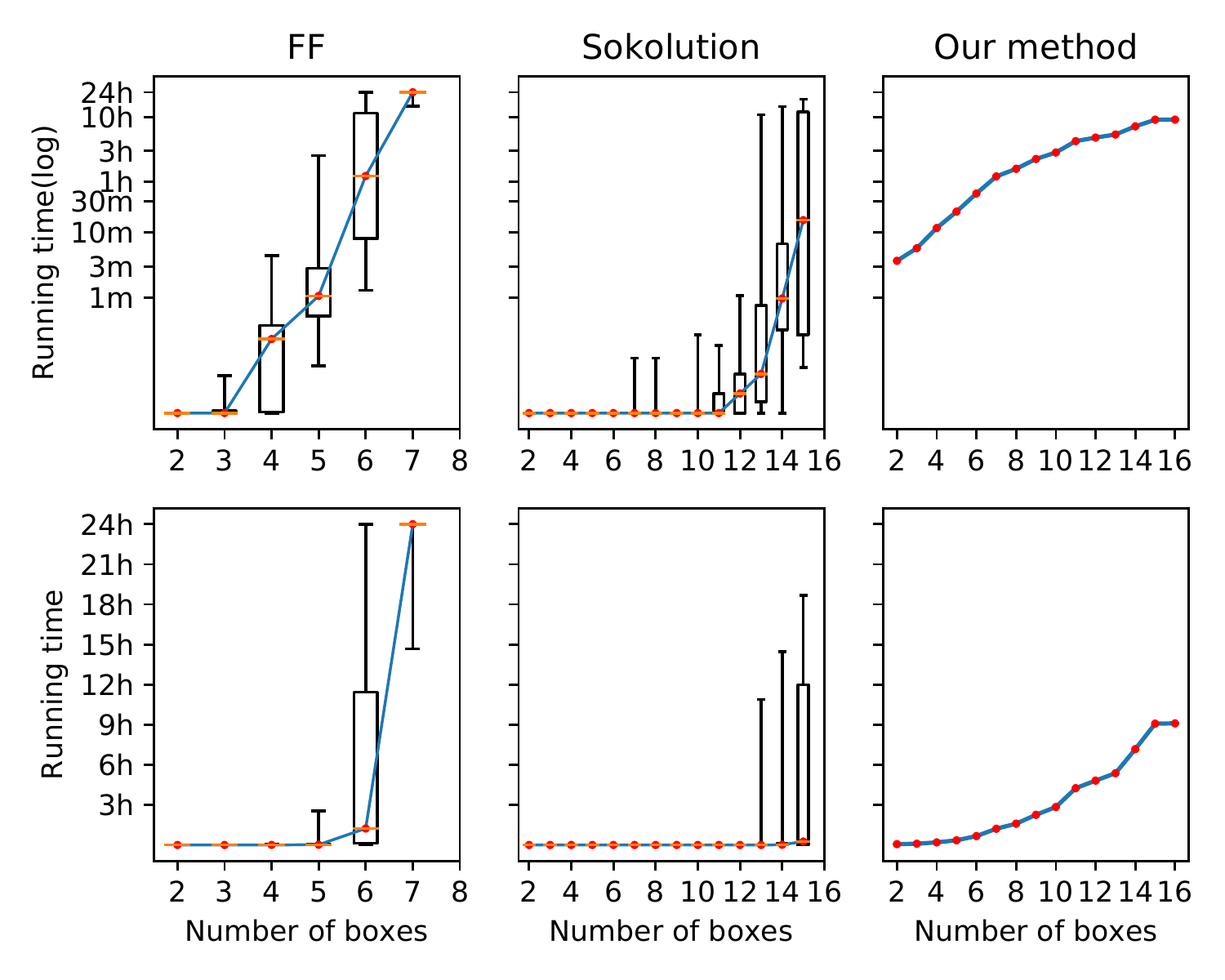}
    \caption{Scaling performance comparison between FF, Sokolution and our method on instance XSokoban\_29. For $m$-box subcases, we randomly generated 19 subcases for FF and Sokolution, and plot their boxplot according to running time. The running time in the top three figures is in logarithmic scale while the bottom three are in linear scale. For our method, we plot the total time needed for the algorithm to achieve $95\%$ success rate for each $m$.
    }
    \label{fig:boxplot}
\end{figure}

Since our framework utilizes extra GPU resources, to gain more insights about the difference between the ways that our framework and traditional search-based algorithms handle hard instances, we evaluate the scaling performance of each solver on subcases with gradually increasing difficulty. We use XSokoban\_29, which contains 16 boxes, for illustration purposes, as shown in Figure \ref{fig:boxplot}. The running time for FF and Sokolution clearly show exponential growth and FF can no longer solve any $m$-box subcase for $m\geq 8$. Sokolution significantly outperforms our method for small-box subcases. We believe this is mainly due to the heavy overhead of the training of neural networks. As the size of subcases increases, our method shines both in running time and scaling performance. Note that our method spends almost no extra time jumping from 15-box subcases to the original 16-box instance. That's because the prediction of the network learned from 15-box subcases is highly accurate on the 16-box instance and the model is already capable of solving the original problem.

\subsection{Exploration Efficiency}
\begin{figure}[t]
    \centering
    \includegraphics[width=1.0\linewidth]{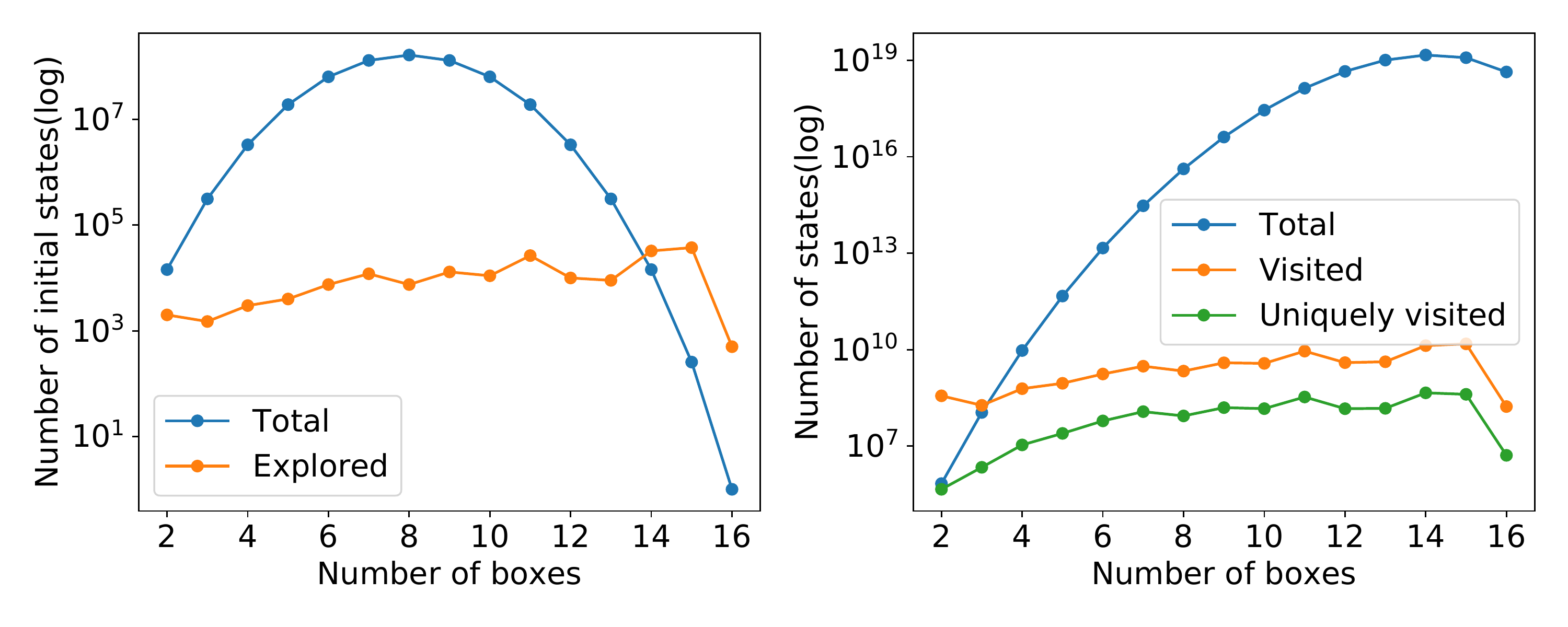}
    \caption{The state statistics during learning. The left figure shows the total number of initial states and the number of explored initial states by the Monte Carlo tree search. The right figure shows the total number of all possible states, all states, and unique states explored by the Monte Carlo tree search. The y-axis of both figures is plotted in logarithmic scale.
    }
    \label{fig:num_state_vs_num_box}
\end{figure}

We now show that the network can efficiently extract knowledge when exploring and generalize to unseen states. For the same XSokoban\_29 instance, we plot the state efficiency by comparing the number of seen states during the exploration phase and the total number of possible states in Figure \ref{fig:num_state_vs_num_box}. In the left figure, for each $m$, the total number of initial states is ${16\choose m}^2$. The number of explored states almost remains at the same magnitude as $m$ increases. This implies the capability of the neural network to efficiently extract structure information of the combinatorial search space and generalize its knowledge to unseen search spaces. The right figure shows the comparison between total possible board states and those explored by the Monte Carlo tree search.

Also notice that for subcases with $m\leq 3$, the model needs to see almost all possible board states before jumping to next stages. This implies generalization does not start for small-box subcases and the model needs to explore every possible combination of board states to understand the underlying structure. As $m$ increases and the combinatorial space grows, generalization starts to shine and help the Monte Carlo tree search stay around the most promising search space.

\subsection{Forgetting during Curriculum Learning}
\begin{figure}[t]
    \centering
    \includegraphics[width=1.0\linewidth]{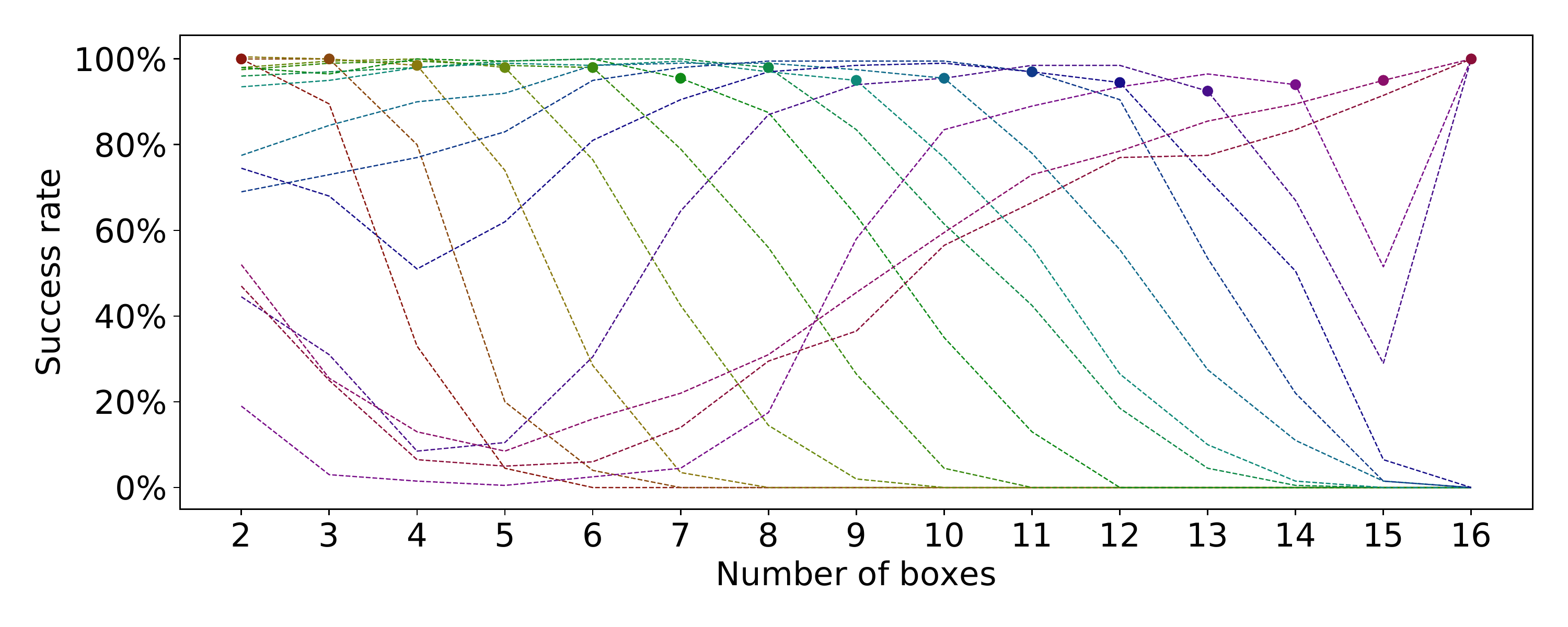}
    \caption{Success rate of different models on different subcases. For each $n \in [2, 16]$, we extracted the model $\mathcal{N}_n$ when the algorithm reached $95\%$ success rate for the first time on $n$-box subcases. Each curve represents a model $\mathcal{N}$, and for each curve there is a corresponding circle on it whose x-coordinate $n$ indicates the model $\mathcal{N}_n$. The x-axis represents each $m$-box subcase, and for each model $\mathcal{N}_n$ we randomly generated $500$ $m$-box subcases and tested its success rate on these subcases.
    }
    \label{fig:sucess_rate}
\end{figure}
One surprising phenomenon in curriculum-driven learning is that the networks may start to forget previously learned structure as the learning proceeds. As seen in Figure \ref{fig:sucess_rate}, as the number of boxes increases, the success rate of small-box subcases gradually decreases. Specifically, we see that the curves trained on higher numbers of boxes drop off to the left, i.e., the performance on cases with fewer boxes becomes worse. On the other hand, the ability to solve increasingly hard cases (more boxes) through the curriculum-driven training shows that knowledge learned from $m$-box subcases can be useful in finding solutions to $m'$-box subcases where $m'>m$. The curve shows that the model learned from $13$-box subcases is already capable of solving the original instance with 16 boxes. This implies that an ensemble of knowledge from small-box subcases can work together to provide enough guidance for finding a solution of the original, unsolved problem instance.

Note that catastrophic forgetting has been previously observed in the context of training deep neural networks. Specifically, deep nets can gradually or abruptly forget previously learned knowledge upon learning new information. This is an important issue to consider because humans typically do not show such catastrophic forgetting when increasing their proficiency on a task. For example, a chess player reaching grand master level will not suddenly start lose to a beginner player. An interesting research challenge is to develop training curricula that prevent catastrophic forgetting for deep RL.

\subsection{Knowledge Extraction from the Network}
\begin{figure}[t]
    \centering
    \includegraphics[width=1.0\linewidth]{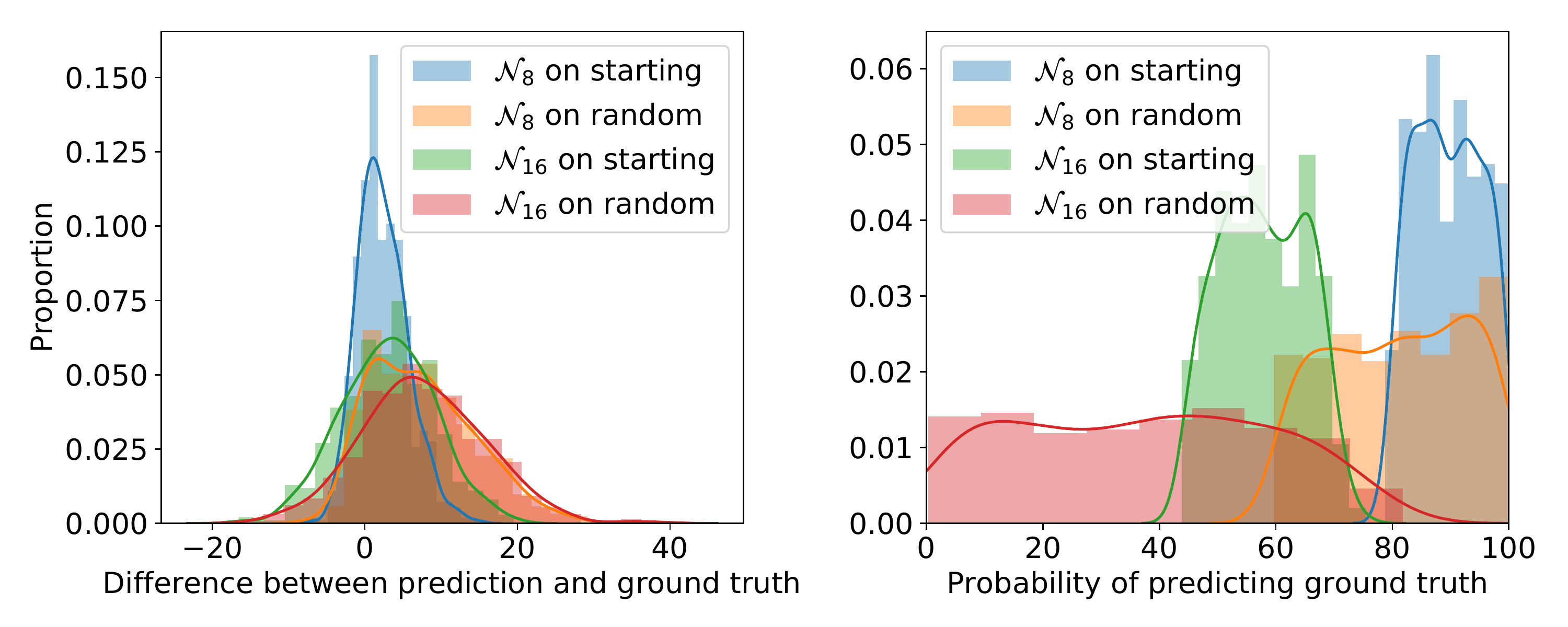}
    \caption{The accuracy of network prediction compared with ground truth. The left figure shows the difference between value prediction which indicates the remaining steps to the goal and the ground truth. The right figures shows the confidence of the network about the ground truth action. We test both on $\mathcal{N}_8$ and $\mathcal{N}_{16}$ that are trained after 8-box subcases and the original problem.
    }
    \label{fig:acc}
\end{figure}
We now show how accurate both value prediction and probability prediction are compared with ground truth provided by optimal Sokoban solvers. We also want to test whether the network can generalize to absolutely unseen board states or just learns well on states that are frequently visited by the Monte Carlo tree search. For this experiment, we test on $m$-box subcases of XSokoban\_29 where $m=8$. We randomly select 500 initial 8-box subcases and do some random pushes on them to generate the set of starting states which are supposed to be frequently seen by the Monte Carlo tree search. And we also generate 500 board states whose boxes are randomly selected from all possible locations of the board. These states are supposed to be hardly explored. All test states are guaranteed to be solvable and dead-end free.

As shown in figure \ref{fig:acc}, we see that the utility function captures the distance to the goal for 8-box subcases surprisingly well for states where the 8 boxes are close to the initial 16-box setup. When we consider subcases with the 8 boxes initially placed randomly, we see the utility function degrade. So the learning does focus on states close to the states that may occur as legal intermediate states which are heavily explored and exploited by reinforcement learning.

When we consider the 16-box learned network, we see an analogous phenomenon but overall less accurate in terms of both utility and policy compared with 8-box scenarios. In fact, the policy for 8-box subcases for randomly placed boxes becomes worst, though still way better than random policy. This means that the Monte Carlo tree search is no longer focused enough to find the goal state in 8-box scenarios. This explains the forgetting curve as discussed earlier.

\section{Conclusion}
% Bart: revised quite a bit for language and summary
We presented a framework based on deep RL for solving hard
combinatorial planning problems in the domain of Sokoban. A key
challenge in the application of deep RL in a single agent setting
is the lack of a positive reinforcement signal since our goal
is to solve  previously unsolved instances that are beyond existing 
combinatorial search methods. We showed how a
curriculum-driven deep RL approach can successfully address this
challenge. By devising a sequence of increasingly complex sub problems, each derived from the original instance, we can incrementally learn an approximate distance to goal function that
can guide MCTS to solving the original problem instance.

We showed the effectiveness of our learning based planning 
strategy by solving hard Sokoban instances that are out of
reach of previous search-based solution techniques, including 
methods specialized for Sokoban. We could uncover plans with
over two hundred actions, where almost any deviation from the plan would lead to an unrecoverable state. Since Sokoban is
one of the hardest challenge domains for current AI planners, 
this work shows the potential of curriculum-based deep RL for solving hard AI planning tasks. In future work, we hope to
extend these techniques to boost theorem proving methods to
find intricate mathematical proofs consisting of long sequences of inference steps to assist in mathematical discovery.

\noindent
\section*{Acknowledgments} 
We  thank  the  anonymous  reviewers for their valuable feedback. This work was supported by the  Center for Human-Compatible AI, an  NSF Expeditions in Computing award (CCF-1522054), an AFSOR award (FA9550-17-1-0292), and an ARO DURIP award (W911NF-17-1-0187) for the compute cluster used in our experimental work. 

% previous text
% a new framework in learning for planning, based on creating a series of curriculum subcases. We use a deep neural network to extract structural features from raw board representation and augment it with Monte Carlo tree search to further enhance its ability of combinatorial search. The network can learn accurate remaining distance prediction and action probability prediction. More importantly, we make solving a hard instance possible by using a curriculum-driven setting, which does not require extra data and handcrafted solutions.

% Our result on challenging Sokoban instance solving suggests that deep reinforcement learning has the capability to find underlying structure of combinatorial search with proper curriculum guidance. Key to our framework is how to decompose the original hard instance, collect useful learning signals from subcases, and ensemble learned knowledge to solve the original instance. This framework can be adapted to solve challenging planning instances in other domains, e.g., open mathematical problems encoded into SAT and first-order logic.

%% The file named.bst is a bibliography style file for BibTeX 0.99c
\bibliographystyle{named}
\bibliography{ijcai20}

\end{document}